\documentclass[
]{ceurart}

\usepackage{listings}
\usepackage{arydshln}
\usepackage{tabularx}
\begin{document}

\copyrightyear{2023}
\copyrightclause{Copyright for this paper by its authors.
  Use permitted under Creative Commons License Attribution 4.0
  International (CC BY 4.0).}

\conference{CLEF 2023: Conference and Labs of the Evaluation Forum, September 18–21, 2023, Thessaloniki, Greece}


\title{PULSAR at MEDIQA-Sum 2023: Large Language Models Augmented by Synthetic Dialogue Convert Patient Dialogues to Medical Records}

\author[1,2]{Viktor Schlegel}
\author[2]{Hao Li}[email=hao.li-2@manchester.ac.uk]
\author[2]{Yuping Wu}[email=yuping.wu@manchester.ac.uk]
\author[1,3]{Anand Subramanian}
\author[1]{Thanh-Tung Nguyen}
\author[1]{Abhinav Ramesh Kashyap}
\author[4]{Daniel Beck}
\author[2]{Xiaojun Zeng}
\author[2]{Riza Theresa Batista-Navarro}
\author[1,3]{Stefan Winkler}
\author[2]{Goran Nenadic}

\address[1]{ASUS Intelligent Cloud Services (AICS), Singapore.}
\address[2]{Dept. of Computer Science, University of Manchester, United Kingdom.}
\address[3]{Dept. of Computer Science, National University of Singapore, Singapore.}
\address[4]{School of Computing and Information Systems, University of Melbourne, Australia.}

\begin{abstract}
This paper describes PULSAR, our system submission at the ImageClef 2023 MediQA-Sum task on summarising patient-doctor dialogues into clinical records. The proposed framework relies on domain-specific pre-training, to produce a specialised language model which is trained on task-specific natural data augmented by synthetic data generated by a black-box LLM. We find limited evidence towards the efficacy of domain-specific pre-training and data augmentation, while scaling up the language model yields the best performance gains. Our approach was ranked second and third among 13 submissions on task B of the challenge.  Our code is available at \url{https://github.com/yuping-wu/PULSAR}.
\end{abstract}


\maketitle

\section{Introduction}
\label{sec:int}
With the recent successes of generative large language models (LLMs) on a variety of tasks~\cite{sanh2022multitask} and domains~\cite{agrawal2022large}, even in the face of data scarcity~\cite{brown2020language}, there is vivid interest in identifying potential application scenarios that could benefit from the power of LLMs. One of the promising domains is healthcare \cite{singhal2022large} as many administrative tasks involve the transformation of textual data. LLM-based approaches that assist hospital staff in repetitive administrative tasks have the potential to improve operational efficiency and documentation quality, optimise revenue streams, reduce cognitive load on healthcare experts, and ultimately result in better and more effective patient care \cite{rajpurkar2022ai}.

A range of different scenarios have been investigated for the suitability of LLM-based assistance, such as summarising patient progress notes as discharge summaries \cite{searle2023discharge} or identifying problems that need treatment during a patient's hospital course \cite{gao2023bionlp}. One of the potential tasks is summarising doctor-patient dialogue as medical records \cite{MEDIQA-Sum2023}. Dialogue summarisation, an established task in the Natural Language Processing (NLP) community, aims to identify salient topics in a multi-turn dialogue \cite{feng2022survey}. State-of-the-art approaches typically formulate the problem as abstractive summarisation, making the task a prime candidate for further investigation of the potential of LLMs in clinical settings. In this scenario, conversations between patients and doctors need to be transformed into (excerpts of) clinical documentation. For example, if a 27 year old female patient mentions that they are experiencing \emph{``Sore throat, runny nose, dry cough and fever 37.5~$^\circ$C''}, the corresponding entry can be the \emph{``Subjective''} section of a medical record excerpt, e.g., \emph{``Patient is a 27 year old female who presents with sore throat, runny nose dry cough and a fever of 37.5~$^\circ$C.''} This documentation is typically performed by the consulting doctor or an attending nurse. Despite bearing potential impact for automation, with clinical staff spending at least 35 minutes of their time every other day on writing such clinical notes \cite{hripcsak2011use}, this task was underexplored by the NLP community, compared to other hospital-related tasks, such as clinical coding \cite{johnson2016mimic,nguyen2023mimic}, or generating radiology reports \cite{monshi2020deep}. More recently, the task has received more attention \cite{DBLP:conf/eacl/AbachaYFL23}, however studies thus far have either focussed on narrow department selections \cite{DBLP:journals/peerjpre/KaziK19, enarvi2020generating}, did not focus on medical documentation generation \cite{yim2021towards}, or have not released their data publicly \cite{joshi2020dr}.

To that end, the ImageClef 2023 MediSum shared task released a collection of dialogues and corresponding clinical notes in an effort to spark interest and advance the state of the art in dialogue as clinical note summarisation \cite{MEDIQA-Sum2023}. The task revolves around three core sub-tasks: \emph{(A)}  identifying the topic of a conversation from a selection of possible medical note sections (i.e., \emph{``Subjective''} in the previous example), \emph{(B)} summarising conversation snippets to appropriate sections in medical records, and, finally, \emph{(C)} summarising full conversations to full medical records. While conversations are synthetic, the corresponding clinical notes are real, doctor-written documentation. 

Our guiding objective to participate in this task was to investigate, how well a recently proposed LLM training framework can generalise to new tasks with as little adaptation as possible \cite{li2023pulsar}. At its core, the framework \emph{(i)} fine-tunes a LLM with a pre-training objective that learns to reconstruct a pseudo-summary consisting of automatically extracted medical terms and \emph{(ii)} employs data augmentation (DA) by instructing Black-Box LLMs to obtain task-specific training data. As such, the DA framework supports any LLM, such as Bloom \cite{scao2022bloom}, GPT-3 \cite{DBLP:conf/nips/BrownMRSKDNSSAA20} or GPT-3.5~\cite{ouyang2022training}.%\footnote{https://chat.openai.com\label{chatgpt}}.

Our submission for task B was ranked second best overall among all participants. Although we have not actively sought to compete in Task C, we observed that our data augmentation technique could improve the performance, particularly when the training data is scarce. These findings underline the potential of LLMs in various settings as well as the generalisability of our proposed approach. %The paper is organised as follows: we describe the challenge in detail in Section~\ref{sec:def}, we introduce our approach in Section~\ref{sec:meth}, we present the results of an empiricla evaluation of our approach in Section~\ref{sec:exp} and we conclude with a discussion of our findings in Section~\ref{sec:disc}.

%-- high level Task Description

%-- Related work

%-- Rationale \& Research Questions

%-- Achievements

\section{Task Definition}
\label{sec:def}
In this section, we describe and formalise the three tasks of the ImageClef 2023 MediSum challenge.
\paragraph{Task A -- Dialogue2Topic Classification} In this task, participants need to identify the topic of a conversation. The list of possible topics corresponds to the 20 different fine-grained sections that can be part of a medical record, such as \emph{``Subjective''}, i.e., the subjective description of symptoms by the patient. %or ``\emph{History of present illness}'' that describes the progression of the illness the patient is treated for. From a machine learning perspective, the task is a multi-class classification. The evaluation metric for this task is accuracy. There are training 1201 and 100 validation examples, respectively. 200 examples form the test set.

% -- Briefly mention Subtask A for completeness

\paragraph{Task B -- Dialogue2Note Summarization} Here, participating systems need to convert a conversation on a specific topic into a corresponding section in the medical record. This task can be regarded as conditional generation, sequence-to-sequence translation or abstractive summarisation. Approaches are evaluated on multiple natural language generation metrics, both based on n-gram overlap, i.e., ROUGE \cite{lin2004rouge}, as well as semantic similarity \cite{sellam2020bleurt,zhang2020bertscore}. 1201 training and 100 validation examples are provided. 200 examples form the test set.
%-- Subtask B

\paragraph{Task C -- Full-Encounter Dialogue2Note Summarization} This task is formulated similarly to Task B, however here, the inputs are full notes and the evaluated systems need to generate medical record outputs for the four general sections \emph{``Subjective''}, \emph{``Objective Exam''}, \emph{``Objective Results''} and \emph{``Assessment and Plan''}. This task features only 67 training and 20 validation examples, with 40 examples reserved for testing. The systems are evaluated based on their output for each of the sections using the ROUGE metrics from Task B; the results are averaged across all sections. An alternative mode of evaluation combines all outputs into one single record and measures the n-gram overlap by means of the ROUGE score.
%-- Subtask C

\paragraph{} The tasks appear to be arranged as a progression, where, given a dialogue, a segmentation and classification model could segment the topics of the conversation (Task A) to be used as input for a Dialogue Snippet Summarisation Model (Task B), the output of which can be arranged as a full medical record (Task C). However, as our goal was to evaluate how well the proposed framework generalises to the tasks with as little adaptation as possible, we decide not to make any task-specific adaptations even if they could provide beneficial given the particular arrangement of the tasks. Thus, we do not rely on any additional information, treat tasks B and C in isolation, and disregard task A for not being a generative task.

\section{Methodology}
\label{sec:meth}
\begin{figure}
    \centering
    \resizebox{0.85\textwidth}{!}{%
    \includegraphics{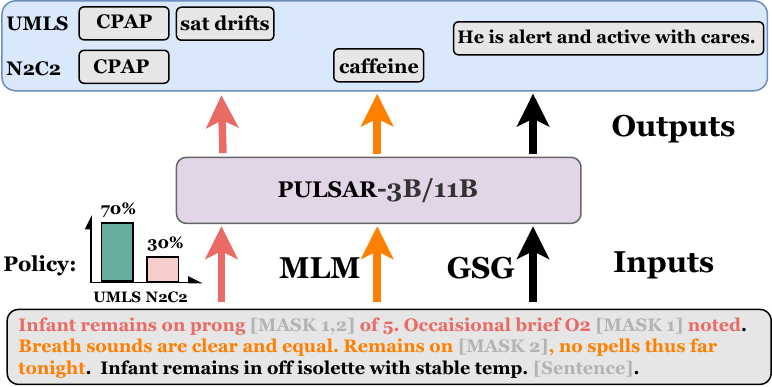}}
    \caption{Example of the pre-train objective of PULSAR. Both Masked Language Model (MLM) and Gap Sentences Generation (GSG) have been employed in this scenario. Red and orange arrows exemplify the UMLS and N2C2 MLM masking strategy, respectively. Meanwhile, the black arrow shows the GSG masking strategy, where a whole sentence has been masked.}
    \label{fig:PULSAR}
\end{figure}

\subsection{Language model Pre-training}
Motivated by the success of predicting masked words \cite{DBLP:journals/tacl/JoshiCLWZL20} and contiguous spans \cite{DBLP:journals/jmlr/RaffelSRLNMZLL20} as self-supervised training objectives, we customised the pre-training objectives for the medical domain generation task to concatenate ``gap text spans (sentences)'' into a pseudo-summary. Each masked span is a medical term from the input text identified by the QuickUMLS \cite{soldaini2016quickumls} or a NER model fine-tuned on a N2C2 dataset (i2b2-2010 challenge \cite{DBLP:journals/jamia/UzunerSSD11}). Specifically, as shown in Figure~\ref{fig:PULSAR}, pre-training consisted of three different policies: first, when both QuickUMLS and N2C2 NER models identified entities, the QuickUMLS results were used in 70\% of cases and the results of the N2C2 NER model were used in 30\%. Second, when only one of them predicted any output, that output was used for masking. Third, when neither had any output, then 15\% of the sentences were masked at random. These text spans were replaced with “sentinel” mask tokens $<extra \underline{~}id\underline{~}i>$ to inform the model that input was masked. In order to provide the model with sufficient medical knowledge, we used the MIMIC-III \cite{johnson2016mimic}, a pre-trained corpus of 2 million data, which consists of a large number of clinical records, such as admission notes, discharge summaries or lab results. %and similar. % 0.1\% of the processed pre-trained corpus did not include tokens from either method (2.2k rows of which did not contain UMLS terms, 23k rows did not contain N2C2-recognised entities, and 797 rows in which neither tool identified any entities).

\subsection{Data Augmentation (DA)}
% rough subsections for now to organize content
Both tasks suffer from scarcity of training data, specifically Task C, which requires generating comprehensive clinical notes based on lengthy patient-doctor conversations based on only 67 training examples. These may be insufficient to train a model capable of performing well on the task. To address this issue, we adopt data augmentation to generate additional examples for training, as this has been shown to improve performance in data-scarce scenarios~\cite{schick2021generating,li2023you}. % As part of PULSAR, we proposed a DA framework that uses a series of black-box LLMs as the backbone. Following previous work, we employ self-debiasing \cite{DBLP:journals/tacl/SchickUS21} to inspire LLMs generation instances that more closely align with instructions i.e. when predicting the next token, not only the probability of the corresponding label is considered, but also the counter label is taken into account. Our task-oriented instructions like ''Given this note, generate a patient doctor conversation'' ...

\paragraph{Prompting Strategy}
 We observed that Large Language Models (LLMs) such as ChatGPT are proficient in understanding clinical context and manipulating clinical data. Therefore, we utilise a pre-existing LLM to generate data for the model's training. Ideally, the data generation approach would involve providing conversations and requesting the LLM to produce the corresponding medical note. However, we are limited by the fact that we only have 67 full-length conversations in our dataset. Nonetheless, we have access to a significantly larger number of medical notes. Hence, we invert the task by prompting the LLM with a medical note (or its snippet) and ask it to generate a hypothetical conversation between the doctor and the patient. We then use the generated conversations as input to train our model to produce the corresponding clinical note.

We employ the OpenAI ChatGPT API (gpt-35-turbo) %\footnote{gpt-35-turbo API} 
for data augmentation, utilising a two-stage prompting strategy to generate data effectively. In the first stage, we use in-context learning with one-shot prompting to prompt the LLM to generate a fictitious conversation between the doctor and patient based on the medical note, while adhering to important guidelines. We provide only one example picked from the training set as we are limited by token context windows for the API. In the second stage (only performed for task C), we prompt the model to include conversational fillers such as ``ums'', ``uh'', and ``hmm'' to the generated conversation from the first stage, as we noticed that the model did not include these fillers despite our instructions in the first stage.

\paragraph{Dataset Utilised}
For task B, we extract matching subsection headings from the MIMIC-III database \cite{johnson2016mimic}, adapting the pre-processing method from \citet{yang2022knowledge} to identify section headers. We rank the generations based on their average Rouge similarity to all training instances and pick the top-scoring $n$ conversations.

For task C, we utilise a corpus of freely available medical notes scraped from MTSamples, which is available on Kaggle\footnote{\url{https://mtsamples.com/} and \url{https://www.kaggle.com/datasets/tboyle10/medicaltranscriptions}, respectively}. Since the dataset contains medical transcriptions of notes from various medical specialities, we devise a method to pick samples from the dataset that are the closest to the medical notes in our training set. To do this, we identify and curate a list of the section headers in the training set through a heuristic approach by exploiting the fact that section headers are usually written in all capital letters. We split the document by newline and extract the lines which are fully upper-cased and add these contents to our list of section headers. We then score the medical notes in MTSamples based on the number of headers that each document has based on the curated list from the previous step and pick the top $n$ documents from MTSamples with the highest scores to use as input for DA. We end up with a corpus of 746 data samples due to the fact that some inputs were flagged as offensive by OpenAI's content moderating policy.

\section{Empirical Evaluation}
\label{sec:exp}
\subsection{Experiments set-up}
We aim to empirically evaluate, how well our framework can solve the problem of converting patient dialogues to medical records. We pursue the following questions:
\begin{enumerate}[(i)]
    \item How well can our proposed approach convert doctor-patient dialogues to Medical Records? 
    \item Does the domain-specific pre-training objective improve performance?
    \item What is the impact of model scale on the performance?
    \item Does synthetic data augmentation improve performance on the tasks?
\end{enumerate}

To answer question \emph{(i)} we empirically evaluate our proposed framework on the task B and C test sets of the ImageClef Challenge. For evidence towards question \emph{(ii)}, we compare the performance of \texttt{PULSAR} to that of equally-sized \texttt{Flan-t5} models. Regarding question \emph{(iii)}, we compare the performance of variously sized models of the same architecture and for question \emph{(iv)}, we compare the performance of models trained on available data only to those fine-tuned on synthetically generated conversation data.
\subsection{Implementation Details}
% submitted models: flan-t5-11b; pulsar-11b (lr 3e-5, epoch 3, train_bs 4, fsdp + CPU offload); llama-13b (lr 3e-4, epoch 3, train_bs 8 + LORA)
\paragraph{Pre-training} \texttt{PULSAR-*} is initialised with weights from the corresponding \texttt{Flan-t5-*} models \cite{chung2022scaling} and pre-trained with four NVIDIA Tesla A100 80GB GPUs for 1 epoch on all MIMIC-III notes. Huggingface Accelerate %\footnote{https://github.com/huggingface/accelerate} 
 is used to optimise GPU memory usage with the Fully Sharded Data Parallel (FSDP) paradigm. We set the training batch size per GPU device as 4 and the gradient accumulation step as 8 to accelerate the training process.

\paragraph{Fine-tuning} We fine-tune all models for 3 epochs. We experiment with encoder-decoder \texttt{Flan-T5}, \texttt{PULSAR} and \texttt{Clinical-T5}~\cite{lehman2023we}%\footnote{https://huggingface.co/xyla/Clinical-T5-Large} 
models, with the configurations \texttt{*-Large} (0.9B Parameters), \texttt{*-3B} and \texttt{*-11B}. Unless stated otherwise, the models are trained on two A100 80GB GPUs with cumulative batch size of 8 and the learning rate of $3^{-5}$. For the largest of them, i.e., (\texttt{Flan-t5-11B} and \texttt{PULSAR-11B}), we use FSDP with CPU offloading. We also experiment with a decoder-only model, \texttt{LLAMA-13B}, freezing and quantising the base model in eight bit~\cite{dettmers2022llm} and using the parameter-efficient LoRA~\cite{hu2022lora} method. More details on hyper-parameter choice are reported in the appendix.

\begin{table}[t!]
\caption{Validation set performance as measured by \{1,2\}-gram overlap Rouge-\{1,2\} and longest sequence overlap Rouge-L Rouge-LSum. Models with the \textsc{*-$n$DG} were augmented with $n$ synthetic examples. The \textsc{*-header} suffix denotes that the section header was used as input.}
\label{tab:overall}
\begin{tabular}{llcccc}

\centering
\textbf{ID} & \textbf{Setting} & \multicolumn{4}{c}{\textbf{Rouge}} \\
& & R1 & R2 & RL & RLSum \\
\hline
& \multicolumn{5}{l}{\emph{\textsc{11B} and \textsc{7B} models}} \\
11B2 &\textsc{PULSAR-11B} & 49.20 & 22.25 & 41.36 & 45.57 \\
11B1 &\textsc{Flan-T5-11B} & 50.75 & 27.92 & 44.93 & 47.58 \\
7B1 &\textsc{LLaMA-7B-LoRA} & 39.3 & 15.7 & 33.1 & 33.1 \\ 
\hdashline
& \multicolumn{5}{l}{\emph{\textsc{3B} models}} \\
3B4 &\textsc{PULSAR-3B-735DG} & 37.89 & 17.83 & 30.40 & 34.76 \\
3B3 &\textsc{Flan-T5-3B-735DG} & 41.44 & 19.05 & 33.93 & 38.41 \\
3B2 &\textsc{PULSAR-3B} & 37.92 & 16.64 & 30.64 & 34.58 \\
3B1 &\textsc{Flan-T5-3B} & 41.91 & 19.41 & 33.76 & 38.04 \\
\hdashline
& \multicolumn{5}{l}{\emph{\textsc{Large} models ($<$ 1B parameters)}} \\
L4 & \textsc{Flan-T5-large-header} & 38.70 & 16.82 & 31.85 & 36.22 \\
L3 & \textsc{Flan-T5-large-1000DG} & 39.41 & 17.04 & 31.85 & 36.78 \\
L2 & \textsc{Flan-T5-large} & 39.27 & 17.42 & 31.95 & 36.49 \\
L1 &\textsc{Clinical-T5-large} & 19.11 & 8.34 & 16.31 & 17.29 \\
\hline
\\
\end{tabular}
\end{table}

\begin{table}[!ht]
\centering
\caption{Test set performance as measured by \{1,2\}-gram overlap Rouge-\{1,2\}, longest sequence overlap Rouge-L and Rouge-LSum and semantic similarity metrics BertScore-F1 and Bleurt; ranked by their aggregation.}
\label{tab:official}
    \begin{tabular}{lccccccc}
        \textbf{submission} & \textbf{R1} & \textbf{R2} & \textbf{RL} & \textbf{RLSum} & \textbf{BS-F1} & \textbf{Bleurt} & \textbf{Agg} \\ \hline
        {SuryaKiran\_run3} & \textbf{43.98} & {18.44} & {35.01} & {35.01} & \textbf{72.31} & \textbf{55.67} & \textbf{57.32} \\
        \textsc{Flan-T5-11B} & 42.99 & \textbf{20.04} & \textbf{35.69} & \textbf{35.69} & 72.18 & 55.49 & 56.89 \\ 
        \textsc{PULSAR-11B} & 41.78 & 19.25 & 34.16 & 34.16 & 72.11 & 55.52 & 56.47 \\ 
        Tredence\_run1 & 42.44 & 17.24 & 35.3 & 35.3 & 72.07 & 53.3 & 55.94 \\ 
        SuryaKiran\_run2 & 42.09 & 18.83 & 34.2 & 34.2 & 71.37 & 54.23 & 55.90 \\ 
        SuryaKiran\_run1 & 40.56 & 17.59 & 32.72 & 32.72 & 71.09 & 53.24 & 54.96 \\ 
        \textsc{LLaMA-7B-LoRA} & 38.15 & 17.3 & 31.42 & 31.42 & 71.77 & 51.52 & 53.82 \\ 
        HuskyScribe\_run1 & 37.67 & 15.04 & 31.26 & 31.26 & 70.54 & 50.37 & 52.86 \\ 
        Tredence\_run2 & 36.21 & 13.84 & 29.66 & 29.66 & 68.82 & 47.29 & 50.77 \\ 
        uetcorn\_run1 & 29.11 & 10.73 & 22.94 & 22.94 & 65.85 & 49.42 & 48.13 \\ 
        uetcorn\_run2 & 28.75 & 10.69 & 23.09 & 23.09 & 65.96 & 49.22 & 47.98 \\ 
        uetcorn\_run3 & 28.72 & 10.7 & 23.04 & 23.04 & 65.92 & 49.13 & 47.92 \\
        SKKU-DSAIL\_run1 & 26.03 & 11.31 & 18.22 & 18.22 & 59.29 & 53.05 & 46.12 \\ 
        
    \hline
    \end{tabular}
\end{table}

\subsection{Results and analysis}
At a glance, Table~\ref{tab:overall} shows the results of our empirical study and Table~\ref{tab:official} shows the final ranking of all participating systems according to the official evaluation by task organisers%\footnote{Other teams' names are pseudonymised as to not unintendedly leak other participants' information.}. 
In the following, we discuss our findings in context of the questions outlined in the motivation of this empirical study.

\paragraph{Our approach generalises well to the dialogue summarisation task.} Overall, our approach generalises well to Task B, with our best model (Table~\ref{tab:overall}, 11B1) surpassing the $50$ Rouge-1 scores mark, which means that on average, half of the prediction tokens are found in reference and vice versa. The high Rouge-L score of $44$ suggests that most of these overlapping tokens indeed form a sequence. However, these scores may be ``boosted'' by the presence of many short target sequences in the evaluation data, such as \emph{``Noncontributory.''} or \emph{``No known allergies.''}, when a dialogue revolves around a topic that does not contribute to the patients' hospital visit.

We find that the utilising the outputs of task A---the section headers---does not contribute to improving the overall performance, compare Table~\ref{tab:overall}, L2 and L4. We observed the same trend across all model sizes (not reported here for brevity).

In the absence of established baselines, we interpret the official rankings of the shared task in Table~\ref{tab:official} as additional evidence towards the success of our approach. 

\paragraph{There is no conclusive evidence that domain-specific pre-training is beneficial.} Comparing 11B1 and 11B2, and 3B1 and 3B2 in Table~\ref{tab:overall}, respectively, we observe that domain-specific pre-training by learning to predict missing medical terms in MIMIC-III notes appears not beneficial, with the gap being smaller for bigger models. One possible reason for this is the domain mismatch between pre-training and application data. MIMIC-III is dominated by inpatient progress notes which track the patients' status along the hospital stay and contain abbreviations, repetitions, incomplete sentences and medical jargon. Conversely, the medical records in the challenge are well-written and stem most likely from admission notes or outpatient encounters, where most of the \emph{initial}  documentation about a new patients' particulars, such as their chief complaint, medical history and drug allergies happens. Additionally, input dialogues have a colloquial tone, further adding to the domain mismatch between pre-training and fine-tuning.

\paragraph{Model scale yields the biggest performance improvements.} Comparing L*, 3B* and 11B* results in Table~\ref{tab:overall}, we can see a clear trend where larger models of the same family consistently perform better. The biggest hike in performance is observed between the 3B and 11B models. This observation is in line with most literature on model scale as driver of performance and the reason for emergent abilities in LLMs \cite{wei2022emergent}. 

We also find that the model trained with adapters can learn to perform on the task successfully, despite the relatively small (around 1.1\% of the full 7B model) number of trainable parameters. However, our results suggest that updating all models' parameters is more effective, as even smaller models outperform the 7B adapter model (Table~\ref{tab:overall}, L2, 3B* compared to 7B1).

\paragraph{Data Augmentation can be helpful if training data is extremely scarce.} Larger models obtain enough signal from the training data of Task B, as there is no clear improvement in scores for the 3B models (Table~\ref{tab:overall}, 3B1 vs. 3B3 and 3B2 vs. 3B4). Meanwhile, data augmentation can lead to consistent, albeit minor, improvements for smaller models (Table~\ref{tab:overall}, L2 vs. L3). When training data is scarce (i.e., Task C) data augmentation helps with the performance. Subjectively, models exhibit typical generation errors such as hallucination and input copying, (see Figure~\ref{fig:examples} in Appendix) and data augmentation seems to alleviate this issue. Quantitatively, data augmentation improves performance across all metrics (27.64 vs 29.41 R1, 9.79 vs 11.60 R2, 16.24 vs 19.18 RL and 23.63 vs 26.08 RLSum without and with DA, respectively). %as shown in Table~\ref{tab:taskc}.
We find the results promising, as the optimised model seems to perform well without any task-specific adaptation. Ultimately, however, this simple approach does not compete with other, potentially task specific information exploiting submissions, with the best of them scoring almost 20 Rouge-1 points higher (20.32 R2, 24.30 RL and 45.06 RLSum).

% \begin{table}
% \centering
% \caption{scores blabla.}
% \label{tab:taskc}
% \begin{tabular}{lcccc}

% \centering
% \textbf{Setting} & \multicolumn{4}{c}{\textbf{Rouge}} \\
% & R1 & R2 & RL & RLSum \\
% \hline
% \textsc{Best} & 49.98 & 20.35 & 24.30 & 45.06 \\
% \hdashline
% \textsc{Flan-T5-3B-746DG} & 29.41 & 11.60 & 19.18 & 26.08 \\
% \textsc{Flan-T5-3B} & 27.64 & 9.79 & 16.24 & 23.62 \\
% \hline
% \\
% \end{tabular}
% \end{table}
\section{Conclusion}
In this work, we present an LLM framework and adapt it to the task of dialogue note summarisation. While we find that the approach generalises well to this new task, there is mixed evidence of the efficacy of both domain-specific pre-training and data augmentation. Our experiments seem to align with the ``bitter lesson of AI''\footnote{\url{http://www.incompleteideas.net/IncIdeas/BitterLesson.html}}, in that model scale seems to trump domain-specific adaptations. This, in turn, supports the narrative of the transformative potential of LLMs in healthcare \cite{qiu2023large}, as larger LLMs become more readily available.

Our findings suggest further avenues for future work: We argued that the pre-training objective may suffer from domain mismatch. As such, experimenting with other domain-specific objectives might improve the performance of the downstream tasks. Furthermore, it is unclear how the choice of hyper-parameters for both training and inference stages (i.e., decoding arguments) impacts the overall performance. Finally, we have left it for future work to investigate, whether data augmentation could provide beneficial with a more advanced filtering strategy, for example by only augmenting examples with certain length or specific section headers. 
As such, we will expand the work reported in this paper by experimenting with different pre-training objectives, performing a more rigorous hyper-parameter optimisation and investigating the impact of data augmentation more closely.

% \section*{Acknowledgements}
% We would like to acknowledge the use of the Computational Shared Facility at The University of Manchester. 
\section*{Limitations}
The results described in this paper should be interpreted within the following context:
\begin{itemize}
    \item The language of the conversations is English. Due to the dominance of English data during pre-training, it is expected that all LLMs that we inspected perform better on English. It is unclear how well the approach will transfer to other languages.
    \item The conversations are synthetic in that they have been written based on existing medical notes, rather than transcribed from real patient-doctor dialogues. While the quality has been evaluated by medical professionals, it is unclear how well the performance would translate to real-world scenarios.
    \item The obtained results should be regarded as preliminary, as robust empirical results such as hyper-parameter optimisation for fine-tuning, pre-training policy selection, exhaustive search for best-performing prompts for data augmentation and strategies for data selection are often impossible given the time constraints of academic challenges and shared tasks.
\end{itemize}
\label{sec:disc}

\bibliography{main}

\begin{thebibliography}{38}
\expandafter\ifx\csname natexlab\endcsname\relax\def\natexlab#1{#1}\fi
\providecommand{\url}[1]{\texttt{#1}}
\providecommand{\href}[2]{#2}
\providecommand{\path}[1]{#1}
\providecommand{\DOIprefix}{doi:}
\providecommand{\ArXivprefix}{arXiv:}
\providecommand{\URLprefix}{URL: }
\providecommand{\Pubmedprefix}{pmid:}
\providecommand{\doi}[1]{\href{http://dx.doi.org/#1}{\path{#1}}}
\providecommand{\Pubmed}[1]{\href{pmid:#1}{\path{#1}}}
\providecommand{\bibinfo}[2]{#2}
\ifx\xfnm\relax \def\xfnm[#1]{\unskip,\space#1}\fi
%Type = Inproceedings
\bibitem[{Sanh et~al.(2022)Sanh, Webson, Raffel, Bach, Sutawika, Alyafeai,
  Chaffin, Stiegler, Raja, Dey et~al.}]{sanh2022multitask}
\bibinfo{author}{V.~Sanh}, \bibinfo{author}{A.~Webson},
  \bibinfo{author}{C.~Raffel}, \bibinfo{author}{S.~Bach},
  \bibinfo{author}{L.~Sutawika}, \bibinfo{author}{Z.~Alyafeai},
  \bibinfo{author}{A.~Chaffin}, \bibinfo{author}{A.~Stiegler},
  \bibinfo{author}{A.~Raja}, \bibinfo{author}{M.~Dey}, et~al.,
\newblock \bibinfo{title}{Multitask prompted training enables zero-shot task
  generalization},
\newblock in: \bibinfo{booktitle}{International Conference on Learning
  Representations 2022}, \bibinfo{year}{2022}.
%Type = Inproceedings
\bibitem[{Agrawal et~al.(2022)Agrawal, Hegselmann, Lang, Kim, and
  Sontag}]{agrawal2022large}
\bibinfo{author}{M.~Agrawal}, \bibinfo{author}{S.~Hegselmann},
  \bibinfo{author}{H.~Lang}, \bibinfo{author}{Y.~Kim},
  \bibinfo{author}{D.~Sontag},
\newblock \bibinfo{title}{Large language models are few-shot clinical
  information extractors},
\newblock in: \bibinfo{booktitle}{Proceedings of the 2022 Conference on
  Empirical Methods in Natural Language Processing}, \bibinfo{year}{2022}, pp.
  \bibinfo{pages}{1998--2022}.
%Type = Article
\bibitem[{Brown et~al.(2020)Brown, Mann, Ryder, Subbiah, Kaplan, Dhariwal,
  Neelakantan, Shyam, Sastry, Askell et~al.}]{brown2020language}
\bibinfo{author}{T.~Brown}, \bibinfo{author}{B.~Mann},
  \bibinfo{author}{N.~Ryder}, \bibinfo{author}{M.~Subbiah},
  \bibinfo{author}{J.~D. Kaplan}, \bibinfo{author}{P.~Dhariwal},
  \bibinfo{author}{A.~Neelakantan}, \bibinfo{author}{P.~Shyam},
  \bibinfo{author}{G.~Sastry}, \bibinfo{author}{A.~Askell}, et~al.,
\newblock \bibinfo{title}{Language models are few-shot learners},
\newblock \bibinfo{journal}{Advances in neural information processing systems}
  \bibinfo{volume}{33} (\bibinfo{year}{2020}) \bibinfo{pages}{1877--1901}.
%Type = Article
\bibitem[{Singhal et~al.(2022)Singhal, Azizi, Tu, Mahdavi, Wei, Chung, Scales,
  Tanwani, Cole-Lewis, Pfohl et~al.}]{singhal2022large}
\bibinfo{author}{K.~Singhal}, \bibinfo{author}{S.~Azizi},
  \bibinfo{author}{T.~Tu}, \bibinfo{author}{S.~S. Mahdavi},
  \bibinfo{author}{J.~Wei}, \bibinfo{author}{H.~W. Chung},
  \bibinfo{author}{N.~Scales}, \bibinfo{author}{A.~Tanwani},
  \bibinfo{author}{H.~Cole-Lewis}, \bibinfo{author}{S.~Pfohl}, et~al.,
\newblock \bibinfo{title}{Large language models encode clinical knowledge},
\newblock \bibinfo{journal}{arXiv preprint arXiv:2212.13138}
  (\bibinfo{year}{2022}).
%Type = Article
\bibitem[{Rajpurkar et~al.(2022)Rajpurkar, Chen, Banerjee, and
  Topol}]{rajpurkar2022ai}
\bibinfo{author}{P.~Rajpurkar}, \bibinfo{author}{E.~Chen},
  \bibinfo{author}{O.~Banerjee}, \bibinfo{author}{E.~J. Topol},
\newblock \bibinfo{title}{Ai in health and medicine},
\newblock \bibinfo{journal}{Nature medicine} \bibinfo{volume}{28}
  (\bibinfo{year}{2022}) \bibinfo{pages}{31--38}.
%Type = Article
\bibitem[{Searle et~al.(2023)Searle, Ibrahim, Teo, and
  Dobson}]{searle2023discharge}
\bibinfo{author}{T.~Searle}, \bibinfo{author}{Z.~Ibrahim},
  \bibinfo{author}{J.~Teo}, \bibinfo{author}{R.~J. Dobson},
\newblock \bibinfo{title}{Discharge summary hospital course summarisation of in
  patient electronic health record text with clinical concept guided deep
  pre-trained transformer models},
\newblock \bibinfo{journal}{Journal of Biomedical Informatics}
  \bibinfo{volume}{141} (\bibinfo{year}{2023}) \bibinfo{pages}{104358}.
%Type = Article
\bibitem[{Gao et~al.(2023)Gao, Miller, Afshar, and Dligach}]{gao2023bionlp}
\bibinfo{author}{Y.~Gao}, \bibinfo{author}{T.~Miller},
  \bibinfo{author}{M.~Afshar}, \bibinfo{author}{D.~Dligach},
\newblock \bibinfo{title}{Bionlp workshop 2023 shared task 1a: Problem list
  summarization},
\newblock \bibinfo{journal}{"Proceedings of the 22nd Workshop on Biomedical
  Language Processing"}  (\bibinfo{year}{2023}).
%Type = Inproceedings
\bibitem[{Yim et~al.(2023)Yim, {Ben Abacha}, Snider, Adams, and
  Yetisgen}]{MEDIQA-Sum2023}
\bibinfo{author}{W.~Yim}, \bibinfo{author}{A.~{Ben Abacha}},
  \bibinfo{author}{N.~Snider}, \bibinfo{author}{G.~Adams},
  \bibinfo{author}{M.~Yetisgen},
\newblock \bibinfo{title}{Overview of the mediqa-sum task at imageclef 2023:
  Summarization and classification of doctor-patient conversations},
\newblock in: \bibinfo{booktitle}{CLEF 2023 Working Notes}, {CEUR} Workshop
  Proceedings, \bibinfo{publisher}{CEUR-WS.org},
  \bibinfo{address}{Thessaloniki, Greece}, \bibinfo{year}{2023}.
%Type = Inproceedings
\bibitem[{Feng et~al.(2022)Feng, Feng, and Qin}]{feng2022survey}
\bibinfo{author}{X.~Feng}, \bibinfo{author}{X.~Feng}, \bibinfo{author}{B.~Qin},
\newblock \bibinfo{title}{A survey on dialogue summarization: Recent advances
  and new frontiers},
\newblock in: \bibinfo{booktitle}{CLEF 2023 Working Notes}, {CEUR} Workshop
  Proceedings, \bibinfo{year}{2022}.
%Type = Article
\bibitem[{Hripcsak et~al.(2011)Hripcsak, Vawdrey, Fred, and
  Bostwick}]{hripcsak2011use}
\bibinfo{author}{G.~Hripcsak}, \bibinfo{author}{D.~K. Vawdrey},
  \bibinfo{author}{M.~R. Fred}, \bibinfo{author}{S.~B. Bostwick},
\newblock \bibinfo{title}{Use of electronic clinical documentation: time spent
  and team interactions},
\newblock \bibinfo{journal}{Journal of the American Medical Informatics
  Association} \bibinfo{volume}{18} (\bibinfo{year}{2011})
  \bibinfo{pages}{112--117}.
%Type = Article
\bibitem[{Johnson et~al.(2016)Johnson, Pollard, Shen, Lehman, Feng, Ghassemi,
  Moody, Szolovits, Anthony~Celi, and Mark}]{johnson2016mimic}
\bibinfo{author}{A.~E. Johnson}, \bibinfo{author}{T.~J. Pollard},
  \bibinfo{author}{L.~Shen}, \bibinfo{author}{L.-w.~H. Lehman},
  \bibinfo{author}{M.~Feng}, \bibinfo{author}{M.~Ghassemi},
  \bibinfo{author}{B.~Moody}, \bibinfo{author}{P.~Szolovits},
  \bibinfo{author}{L.~Anthony~Celi}, \bibinfo{author}{R.~G. Mark},
\newblock \bibinfo{title}{Mimic-iii, a freely accessible critical care
  database},
\newblock \bibinfo{journal}{Scientific data} \bibinfo{volume}{3}
  (\bibinfo{year}{2016}) \bibinfo{pages}{1--9}.
%Type = Article
\bibitem[{Nguyen et~al.(2023)Nguyen, Schlegel, Kashyap, Winkler, Huang, Liu,
  and Lin}]{nguyen2023mimic}
\bibinfo{author}{T.-T. Nguyen}, \bibinfo{author}{V.~Schlegel},
  \bibinfo{author}{A.~Kashyap}, \bibinfo{author}{S.~Winkler},
  \bibinfo{author}{S.-S. Huang}, \bibinfo{author}{J.-J. Liu},
  \bibinfo{author}{C.-J. Lin},
\newblock \bibinfo{title}{Mimic-iv-icd: A new benchmark for extreme multilabel
  classification},
\newblock \bibinfo{journal}{arXiv preprint arXiv:2304.13998}
  (\bibinfo{year}{2023}).
%Type = Article
\bibitem[{Monshi et~al.(2020)Monshi, Poon, and Chung}]{monshi2020deep}
\bibinfo{author}{M.~M.~A. Monshi}, \bibinfo{author}{J.~Poon},
  \bibinfo{author}{V.~Chung},
\newblock \bibinfo{title}{Deep learning in generating radiology reports: A
  survey},
\newblock \bibinfo{journal}{Artificial Intelligence in Medicine}
  \bibinfo{volume}{106} (\bibinfo{year}{2020}) \bibinfo{pages}{101878}.
%Type = Inproceedings
\bibitem[{Abacha et~al.(2023)Abacha, Yim, Fan, and
  Lin}]{DBLP:conf/eacl/AbachaYFL23}
\bibinfo{author}{A.~B. Abacha}, \bibinfo{author}{W.~Yim},
  \bibinfo{author}{Y.~Fan}, \bibinfo{author}{T.~Lin},
\newblock \bibinfo{title}{An empirical study of clinical note generation from
  doctor-patient encounters},
\newblock in: \bibinfo{booktitle}{{EACL}}, \bibinfo{publisher}{Association for
  Computational Linguistics}, \bibinfo{year}{2023}, pp.
  \bibinfo{pages}{2283--2294}.
%Type = Article
\bibitem[{Kazi and Kahanda(2019)}]{DBLP:journals/peerjpre/KaziK19}
\bibinfo{author}{N.~Kazi}, \bibinfo{author}{I.~Kahanda},
\newblock \bibinfo{title}{Automatically generating psychiatric case notes from
  digital transcripts of doctor-patient conversations using text mining},
\newblock \bibinfo{journal}{PeerJ Prepr.} \bibinfo{volume}{7}
  (\bibinfo{year}{2019}) \bibinfo{pages}{e27497}.
%Type = Inproceedings
\bibitem[{Enarvi et~al.(2020)Enarvi, Amoia, Teba, Delaney, Diehl, Hahn, Harris,
  McGrath, Pan, Pinto et~al.}]{enarvi2020generating}
\bibinfo{author}{S.~Enarvi}, \bibinfo{author}{M.~Amoia},
  \bibinfo{author}{M.~D.-A. Teba}, \bibinfo{author}{B.~Delaney},
  \bibinfo{author}{F.~Diehl}, \bibinfo{author}{S.~Hahn},
  \bibinfo{author}{K.~Harris}, \bibinfo{author}{L.~McGrath},
  \bibinfo{author}{Y.~Pan}, \bibinfo{author}{J.~Pinto}, et~al.,
\newblock \bibinfo{title}{Generating medical reports from patient-doctor
  conversations using sequence-to-sequence models},
\newblock in: \bibinfo{booktitle}{Proceedings of the First Workshop on Natural
  Language Processing for Medical Conversations}, \bibinfo{year}{2020}, pp.
  \bibinfo{pages}{22--30}.
%Type = Inproceedings
\bibitem[{Yim and Yetisgen-Yildiz(2021)}]{yim2021towards}
\bibinfo{author}{W.-w. Yim}, \bibinfo{author}{M.~Yetisgen-Yildiz},
\newblock \bibinfo{title}{Towards automating medical scribing: Clinic visit
  dialogue2note sentence alignment and snippet summarization},
\newblock in: \bibinfo{booktitle}{Proceedings of the Second Workshop on Natural
  Language Processing for Medical Conversations}, \bibinfo{year}{2021}, pp.
  \bibinfo{pages}{10--20}.
%Type = Article
\bibitem[{Joshi et~al.(2020)Joshi, Katariya, Amatriain, and
  Kannan}]{joshi2020dr}
\bibinfo{author}{A.~Joshi}, \bibinfo{author}{N.~Katariya},
  \bibinfo{author}{X.~Amatriain}, \bibinfo{author}{A.~Kannan},
\newblock \bibinfo{title}{Dr. summarize: Global summarization of medical
  dialogue by exploiting local structures},
\newblock \bibinfo{journal}{arXiv preprint arXiv:2009.08666}
  (\bibinfo{year}{2020}).
%Type = Article
\bibitem[{Li et~al.(2023)Li, Wu, Schlegel, Batista-Navarro, Nguyen,
  Ramesh~Kashyap, Zeng, Beck, Winkler, and Nenadic}]{li2023pulsar}
\bibinfo{author}{H.~Li}, \bibinfo{author}{Y.~Wu},
  \bibinfo{author}{V.~Schlegel}, \bibinfo{author}{R.~Batista-Navarro},
  \bibinfo{author}{T.-T. Nguyen}, \bibinfo{author}{A.~Ramesh~Kashyap},
  \bibinfo{author}{X.~Zeng}, \bibinfo{author}{D.~Beck},
  \bibinfo{author}{S.~Winkler}, \bibinfo{author}{G.~Nenadic},
\newblock \bibinfo{title}{Pulsar: Pre-training with extracted healthcare terms
  for summarising patients’ problems and data augmentation with black-box
  large language models},
\newblock \bibinfo{journal}{arXiv preprint}  (\bibinfo{year}{2023}).
%Type = Article
\bibitem[{Scao et~al.(2022)Scao, Fan, Akiki, Pavlick, Ili{\'c}, Hesslow,
  Castagn{\'e}, Luccioni, Yvon, Gall{\'e} et~al.}]{scao2022bloom}
\bibinfo{author}{T.~L. Scao}, \bibinfo{author}{A.~Fan},
  \bibinfo{author}{C.~Akiki}, \bibinfo{author}{E.~Pavlick},
  \bibinfo{author}{S.~Ili{\'c}}, \bibinfo{author}{D.~Hesslow},
  \bibinfo{author}{R.~Castagn{\'e}}, \bibinfo{author}{A.~S. Luccioni},
  \bibinfo{author}{F.~Yvon}, \bibinfo{author}{M.~Gall{\'e}}, et~al.,
\newblock \bibinfo{title}{Bloom: A 176b-parameter open-access multilingual
  language model},
\newblock \bibinfo{journal}{arXiv preprint arXiv:2211.05100}
  (\bibinfo{year}{2022}).
%Type = Inproceedings
\bibitem[{Brown et~al.(2020)Brown, Mann, Ryder, Subbiah, Kaplan, Dhariwal,
  Neelakantan, Shyam, Sastry, Askell, Agarwal, Herbert{-}Voss, Krueger,
  Henighan, Child, Ramesh, Ziegler, Wu, Winter, Hesse, Chen, Sigler, Litwin,
  Gray, Chess, Clark, Berner, McCandlish, Radford, Sutskever, and
  Amodei}]{DBLP:conf/nips/BrownMRSKDNSSAA20}
\bibinfo{author}{T.~B. Brown}, \bibinfo{author}{B.~Mann},
  \bibinfo{author}{N.~Ryder}, \bibinfo{author}{M.~Subbiah},
  \bibinfo{author}{J.~Kaplan}, \bibinfo{author}{P.~Dhariwal},
  \bibinfo{author}{A.~Neelakantan}, \bibinfo{author}{P.~Shyam},
  \bibinfo{author}{G.~Sastry}, \bibinfo{author}{A.~Askell},
  \bibinfo{author}{S.~Agarwal}, \bibinfo{author}{A.~Herbert{-}Voss},
  \bibinfo{author}{G.~Krueger}, \bibinfo{author}{T.~Henighan},
  \bibinfo{author}{R.~Child}, \bibinfo{author}{A.~Ramesh},
  \bibinfo{author}{D.~M. Ziegler}, \bibinfo{author}{J.~Wu},
  \bibinfo{author}{C.~Winter}, \bibinfo{author}{C.~Hesse},
  \bibinfo{author}{M.~Chen}, \bibinfo{author}{E.~Sigler},
  \bibinfo{author}{M.~Litwin}, \bibinfo{author}{S.~Gray},
  \bibinfo{author}{B.~Chess}, \bibinfo{author}{J.~Clark},
  \bibinfo{author}{C.~Berner}, \bibinfo{author}{S.~McCandlish},
  \bibinfo{author}{A.~Radford}, \bibinfo{author}{I.~Sutskever},
  \bibinfo{author}{D.~Amodei},
\newblock \bibinfo{title}{Language models are few-shot learners},
\newblock in: \bibinfo{booktitle}{NeurIPS}, \bibinfo{year}{2020}.
%Type = Inproceedings
\bibitem[{Ouyang et~al.(2022)Ouyang, Wu, Jiang, Almeida, Wainwright, Mishkin,
  Zhang, Agarwal, Slama, Ray et~al.}]{ouyang2022training}
\bibinfo{author}{L.~Ouyang}, \bibinfo{author}{J.~Wu},
  \bibinfo{author}{X.~Jiang}, \bibinfo{author}{D.~Almeida},
  \bibinfo{author}{C.~Wainwright}, \bibinfo{author}{P.~Mishkin},
  \bibinfo{author}{C.~Zhang}, \bibinfo{author}{S.~Agarwal},
  \bibinfo{author}{K.~Slama}, \bibinfo{author}{A.~Ray}, et~al.,
\newblock \bibinfo{title}{Training language models to follow instructions with
  human feedback},
\newblock in: \bibinfo{booktitle}{Advances in Neural Information Processing
  Systems}, volume~\bibinfo{volume}{35}, \bibinfo{year}{2022}, pp.
  \bibinfo{pages}{27730--27744}.
%Type = Inproceedings
\bibitem[{Lin(2004)}]{lin2004rouge}
\bibinfo{author}{C.-Y. Lin},
\newblock \bibinfo{title}{Rouge: A package for automatic evaluation of
  summaries},
\newblock in: \bibinfo{booktitle}{Text summarization branches out},
  \bibinfo{year}{2004}, pp. \bibinfo{pages}{74--81}.
%Type = Inproceedings
\bibitem[{Sellam et~al.(2020)Sellam, Das, and Parikh}]{sellam2020bleurt}
\bibinfo{author}{T.~Sellam}, \bibinfo{author}{D.~Das},
  \bibinfo{author}{A.~Parikh},
\newblock \bibinfo{title}{Bleurt: Learning robust metrics for text generation},
\newblock in: \bibinfo{booktitle}{Proceedings of the 58th Annual Meeting of the
  Association for Computational Linguistics}, \bibinfo{year}{2020}, pp.
  \bibinfo{pages}{7881--7892}.
%Type = Inproceedings
\bibitem[{Zhang et~al.(2020)Zhang, Kishore, Wu, Weinberger, and
  Artzi}]{zhang2020bertscore}
\bibinfo{author}{T.~Zhang}, \bibinfo{author}{V.~Kishore},
  \bibinfo{author}{F.~Wu}, \bibinfo{author}{K.~Q. Weinberger},
  \bibinfo{author}{Y.~Artzi},
\newblock \bibinfo{title}{Bertscore: Evaluating text generation with bert},
\newblock in: \bibinfo{booktitle}{International Conference on Learning
  Representations}, \bibinfo{year}{2020}.
%Type = Article
\bibitem[{Joshi et~al.(2020)Joshi, Chen, Liu, Weld, Zettlemoyer, and
  Levy}]{DBLP:journals/tacl/JoshiCLWZL20}
\bibinfo{author}{M.~Joshi}, \bibinfo{author}{D.~Chen},
  \bibinfo{author}{Y.~Liu}, \bibinfo{author}{D.~S. Weld},
  \bibinfo{author}{L.~Zettlemoyer}, \bibinfo{author}{O.~Levy},
\newblock \bibinfo{title}{Spanbert: Improving pre-training by representing and
  predicting spans},
\newblock \bibinfo{journal}{Trans. Assoc. Comput. Linguistics}
  \bibinfo{volume}{8} (\bibinfo{year}{2020}) \bibinfo{pages}{64--77}.
%Type = Article
\bibitem[{Raffel et~al.(2020)Raffel, Shazeer, Roberts, Lee, Narang, Matena,
  Zhou, Li, and Liu}]{DBLP:journals/jmlr/RaffelSRLNMZLL20}
\bibinfo{author}{C.~Raffel}, \bibinfo{author}{N.~Shazeer},
  \bibinfo{author}{A.~Roberts}, \bibinfo{author}{K.~Lee},
  \bibinfo{author}{S.~Narang}, \bibinfo{author}{M.~Matena},
  \bibinfo{author}{Y.~Zhou}, \bibinfo{author}{W.~Li}, \bibinfo{author}{P.~J.
  Liu},
\newblock \bibinfo{title}{Exploring the limits of transfer learning with a
  unified text-to-text transformer},
\newblock \bibinfo{journal}{J. Mach. Learn. Res.} \bibinfo{volume}{21}
  (\bibinfo{year}{2020}) \bibinfo{pages}{140:1--140:67}.
%Type = Inproceedings
\bibitem[{Soldaini and Goharian(2016)}]{soldaini2016quickumls}
\bibinfo{author}{L.~Soldaini}, \bibinfo{author}{N.~Goharian},
\newblock \bibinfo{title}{Quickumls: a fast, unsupervised approach for medical
  concept extraction},
\newblock in: \bibinfo{booktitle}{MedIR workshop, sigir}, \bibinfo{year}{2016},
  pp. \bibinfo{pages}{1--4}.
%Type = Article
\bibitem[{Uzuner et~al.(2011)Uzuner, South, Shen, and
  DuVall}]{DBLP:journals/jamia/UzunerSSD11}
\bibinfo{author}{{\"{O}}.~Uzuner}, \bibinfo{author}{B.~R. South},
  \bibinfo{author}{S.~Shen}, \bibinfo{author}{S.~L. DuVall},
\newblock \bibinfo{title}{2010 i2b2/va challenge on concepts, assertions, and
  relations in clinical text},
\newblock \bibinfo{journal}{J. Am. Medical Informatics Assoc.}
  \bibinfo{volume}{18} (\bibinfo{year}{2011}) \bibinfo{pages}{552--556}.
%Type = Inproceedings
\bibitem[{Schick and Sch{\"u}tze(2021)}]{schick2021generating}
\bibinfo{author}{T.~Schick}, \bibinfo{author}{H.~Sch{\"u}tze},
\newblock \bibinfo{title}{Generating datasets with pretrained language models},
\newblock in: \bibinfo{booktitle}{Proceedings of the 2021 Conference on
  Empirical Methods in Natural Language Processing}, \bibinfo{year}{2021}, pp.
  \bibinfo{pages}{6943--6951}.
%Type = Article
\bibitem[{Li et~al.(2023)Li, Schlegel, Batista-Navarro, and
  Nenadic}]{li2023you}
\bibinfo{author}{H.~Li}, \bibinfo{author}{V.~Schlegel},
  \bibinfo{author}{R.~Batista-Navarro}, \bibinfo{author}{G.~Nenadic},
\newblock \bibinfo{title}{Do you hear the people sing? key point analysis via
  iterative clustering and abstractive summarisation},
\newblock \bibinfo{journal}{arXiv preprint arXiv:2305.16000}
  (\bibinfo{year}{2023}).
%Type = Article
\bibitem[{Yang et~al.(2022)Yang, Wang, Rawat, Mitra, and
  Yu}]{yang2022knowledge}
\bibinfo{author}{Z.~Yang}, \bibinfo{author}{S.~Wang}, \bibinfo{author}{B.~P.~S.
  Rawat}, \bibinfo{author}{A.~Mitra}, \bibinfo{author}{H.~Yu},
\newblock \bibinfo{title}{Knowledge injected prompt based fine-tuning for
  multi-label few-shot icd coding},
\newblock \bibinfo{journal}{arXiv preprint arXiv:2210.03304}
  (\bibinfo{year}{2022}).
%Type = Article
\bibitem[{Chung et~al.(2022)Chung, Hou, Longpre, Zoph, Tay, Fedus, Li, Wang,
  Dehghani, Brahma et~al.}]{chung2022scaling}
\bibinfo{author}{H.~W. Chung}, \bibinfo{author}{L.~Hou},
  \bibinfo{author}{S.~Longpre}, \bibinfo{author}{B.~Zoph},
  \bibinfo{author}{Y.~Tay}, \bibinfo{author}{W.~Fedus},
  \bibinfo{author}{E.~Li}, \bibinfo{author}{X.~Wang},
  \bibinfo{author}{M.~Dehghani}, \bibinfo{author}{S.~Brahma}, et~al.,
\newblock \bibinfo{title}{Scaling instruction-finetuned language models},
\newblock \bibinfo{journal}{arXiv preprint arXiv:2210.11416}
  (\bibinfo{year}{2022}).
%Type = Article
\bibitem[{Lehman et~al.(2023)Lehman, Hernandez, Mahajan, Wulff, Smith, Ziegler,
  Nadler, Szolovits, Johnson, and Alsentzer}]{lehman2023we}
\bibinfo{author}{E.~Lehman}, \bibinfo{author}{E.~Hernandez},
  \bibinfo{author}{D.~Mahajan}, \bibinfo{author}{J.~Wulff},
  \bibinfo{author}{M.~J. Smith}, \bibinfo{author}{Z.~Ziegler},
  \bibinfo{author}{D.~Nadler}, \bibinfo{author}{P.~Szolovits},
  \bibinfo{author}{A.~Johnson}, \bibinfo{author}{E.~Alsentzer},
\newblock \bibinfo{title}{Do we still need clinical language models?},
\newblock \bibinfo{journal}{arXiv preprint arXiv:2302.08091}
  (\bibinfo{year}{2023}).
%Type = Article
\bibitem[{Dettmers et~al.(2022)Dettmers, Lewis, Belkada, and
  Zettlemoyer}]{dettmers2022llm}
\bibinfo{author}{T.~Dettmers}, \bibinfo{author}{M.~Lewis},
  \bibinfo{author}{Y.~Belkada}, \bibinfo{author}{L.~Zettlemoyer},
\newblock \bibinfo{title}{Llm. int8 (): 8-bit matrix multiplication for
  transformers at scale},
\newblock \bibinfo{journal}{arXiv preprint arXiv:2208.07339}
  (\bibinfo{year}{2022}).
%Type = Inproceedings
\bibitem[{Hu et~al.(2022)Hu, Wallis, Allen-Zhu, Li, Wang, Wang, Chen
  et~al.}]{hu2022lora}
\bibinfo{author}{E.~J. Hu}, \bibinfo{author}{P.~Wallis},
  \bibinfo{author}{Z.~Allen-Zhu}, \bibinfo{author}{Y.~Li},
  \bibinfo{author}{S.~Wang}, \bibinfo{author}{L.~Wang},
  \bibinfo{author}{W.~Chen}, et~al.,
\newblock \bibinfo{title}{Lora: Low-rank adaptation of large language models},
\newblock in: \bibinfo{booktitle}{International Conference on Learning
  Representations}, \bibinfo{year}{2022}.
%Type = Article
\bibitem[{Wei et~al.(2011)Wei, Tay, Bommasani, Raffel, Zoph, Borgeaud,
  Yogatama, Bosma, Zhou, Metzler et~al.}]{wei2022emergent}
\bibinfo{author}{J.~Wei}, \bibinfo{author}{Y.~Tay},
  \bibinfo{author}{R.~Bommasani}, \bibinfo{author}{C.~Raffel},
  \bibinfo{author}{B.~Zoph}, \bibinfo{author}{S.~Borgeaud},
  \bibinfo{author}{D.~Yogatama}, \bibinfo{author}{M.~Bosma},
  \bibinfo{author}{D.~Zhou}, \bibinfo{author}{D.~Metzler}, et~al.,
\newblock \bibinfo{title}{Emergent abilities of large language models},
\newblock \bibinfo{journal}{Transactions on Machine Learning Research}
  \bibinfo{volume}{8} (\bibinfo{year}{2011}).
%Type = Article
\bibitem[{Qiu et~al.(2023)Qiu, Li, Sun, Peng, Shi, Zhang, Dong, Lam, Lo, Xiao
  et~al.}]{qiu2023large}
\bibinfo{author}{J.~Qiu}, \bibinfo{author}{L.~Li}, \bibinfo{author}{J.~Sun},
  \bibinfo{author}{J.~Peng}, \bibinfo{author}{P.~Shi},
  \bibinfo{author}{R.~Zhang}, \bibinfo{author}{Y.~Dong},
  \bibinfo{author}{K.~Lam}, \bibinfo{author}{F.~P.-W. Lo},
  \bibinfo{author}{B.~Xiao}, et~al.,
\newblock \bibinfo{title}{Large ai models in health informatics: Applications,
  challenges, and the future},
\newblock \bibinfo{journal}{arXiv preprint arXiv:2303.11568}
  (\bibinfo{year}{2023}).

\end{thebibliography}

\appendix
\section{Hyper-parameters for training and inference}
We initialise LoRA with $r=16$, $\alpha=16$ on the query, key, value and output projection weights of all layers of the base model ($Q, K, V$ and $O$, respectively). The model is trained on a single A100 80GB GPU with a learning rate of $3^{-4}$ for the adapter weights. For both encoder-decoder and decoder only settings, during training, we optimise the parameters of the language models to minimise the cross-entropy loss between each token of the prediction and the corresponding token of the ground truth answer sequence using teacher forcing. For encoder-decoder models, we limit the length of input dialogues to at most 496 and the length of output notes to at most 214 tokens, respectively (95th percentile). For the decoder model, we limit the length of input and output combined to at most 696 tokens. During inference, we set no limits to input and output sequence lengths and decode the prediction using beam search with 6 (4 for LLaMa), temperature of 1.0, top k of 50 (40 for LLaMa) and top p of 1.0 (0.7 for LLaMa).

For task C, we use the same arguments as for task C, with the exception of limiting the input length to 2048 and output length to 990 during training, in order to fit the GPU during training.

\section{Qualitative example}

Figure~\ref{fig:PULSAR} shows qualitative examples generated by our models trained on task C training data with and without data augmentation, respectively.
\begin{figure}[h!]
    \centering
    \begin{tabularx}{0.95\textwidth}{X|X}
        \textsc{Flan-T5-3B} & \textsc{Flan-T5-3B-746DG} \\
        \hline
         Alexander is a 62-year-old male, with a past medical history significant for reflux, who presents for follow-up of his chronic problems. He is so thankful you put him on that medicine for his reflux. The protonix that he had, w- made an amazing change in his life. He's really comfortable now. He eats whatever he wants, and he feels so much better. The doctor is glad to hear that. Okay. How are you doing, kind of, managing your diet? I know, you have to do some lifestyle modifications, like cutting back on caffeine and spicy foods and alcohol. How are you doing with that? \emph{[patient] I'm doing really well. I moved over from caffeine, over to green tea. [doctor] Okay. [patient] and it, it is so, m- it doesn't cause as much problem as it did with, when he was drinking so many energy drinks a day...}
         & 
        HISTORY OF PRESENT ILLNESS: Alexander is a 62-year-old male with a past medical history significant for reflux. He presents for follow-up of his chronic problems. He has a good support system at home. He has a big family. All his kids call and check on him every day. He does not have any symptoms of chest pain, shortness of breath, belly pain, nausea or vomiting. He does not have any symptoms of nausea or vomiting. On physical examination, I do not hear any carotid bruits in his neck. I do hear a slight 2/6 systolic ejection murmur, which I've heard in the past, so that's stable. His lungs are nice and clear, and he does have 1+ pitting edema bilaterally in his lower extremities. I think he is doing a good job watching his diet. He could just be retaining a little bit of fluid, maybe just from standing all day.
 \\
        \hline
    \end{tabularx}
    \caption{Example excerpts of outputs for Task C produced by models with and without data augmentation. An instance of input copying is highlighted in italics.}
    \label{fig:examples}
\end{figure}
\end{document}